\documentclass[letterpaper, 10 pt, conference]{ieeeconf}  
\let\labelindent\relax

\usepackage[T1]{fontenc}    
\usepackage{url}            
\usepackage{booktabs}       
\usepackage{amsfonts}       
\usepackage{nicefrac}       
\usepackage{microtype}      
\usepackage{multirow}
\usepackage{enumitem}
\usepackage{amsmath,amsfonts}
\usepackage{bm}
\usepackage{graphicx}
\usepackage{listings}
\usepackage{textcomp}
\usepackage{xcolor}
\usepackage{colortbl}
\usepackage{tcolorbox}
\usepackage{xspace}
\usepackage{flushend}
\usepackage{pifont}
\usepackage{wrapfig}
\usepackage{mdframed}
\usepackage{needspace}
\usepackage{hyperref} 
\usepackage{subcaption}
\usepackage{cite}

\makeatletter
\DeclareRobustCommand\onedot{\futurelet\@let@token\@onedot}
\def\@onedot{\ifx\@let@token.\else.\null\fi\xspace}

\makeatother

\newcommand{\responseline}[1]{\textcolor{black}{#1}}

\IEEEoverridecommandlockouts                              

\title{\Large \bf LADEV: A Language-Driven Testing and Evaluation Platform for Vision-Language-Action Models in Robotic Manipulation}

\author{Zhijie Wang$^{1}$, Zhehua Zhou$^{1}$, Jiayang Song$^{1}$, Yuheng Huang$^{2}$, Zhan Shu$^{1}$, and Lei Ma$^{2,1}$
    \thanks{$^{1}$Zhijie Wang, Zhehua Zhou, Jiayang Song, and Zhan Shu are with the University of Alberta, Edmonton, AB, Canada {\tt\small \{zhijie.wang, zhehua1, jiayan13, zshu1\}@ualberta.ca}}%
    \thanks{$^{2}$Yuheng Huang and Lei Ma are with The University of Tokyo, Tokyo, Japan. Lei Ma is also with the University of Alberta {\tt\small yuhenghuang42@g.ecc.u-tokyo.ac.jp, ma.lei@acm.org}.
    The code of this paper will be made available online soon.
    Additional information and materials are provided in~\href{https://sites.google.com/view/ladev}{https://sites.google.com/view/ladev}.}%
}

\definecolor{commentred}{rgb}{0.6, 0.2, 0.2}

\definecolor{lightgray}{gray}{0.85}

\begin{document}

\maketitle
\thispagestyle{empty}
\pagestyle{empty}

\begin{abstract}
Building on the advancements of Large Language Models (LLMs) and Vision Language Models (VLMs), recent research has introduced Vision-Language-Action (VLA) models as an integrated solution for robotic manipulation tasks.
These models take camera images and natural language task instructions as input and directly generate control actions for robots to perform specified tasks, greatly improving both decision-making capabilities and interaction with human users. 
However, the data-driven nature of VLA models, combined with their lack of interpretability, makes the assurance of their effectiveness and robustness a challenging task. 
This highlights the need for a reliable testing and evaluation platform.
For this purpose, in this work, we propose LADEV, a comprehensive and efficient platform specifically designed for evaluating VLA models. 
We first present a language-driven approach that automatically generates simulation environments from natural language inputs, mitigating the need for manual adjustments and significantly improving testing efficiency. 
Then, to further assess the influence of language input to the VLA models, we implement a paraphrase mechanism that produces diverse natural language task instructions for testing. 
Finally, to expedite the evaluation process, we introduce a batch-style method for conducting large-scale testing of VLA models.
Using LADEV, we conducted experiments on several state-of-the-art VLA models, demonstrating its effectiveness as a tool for evaluating these models. 
Our results showed that LADEV not only enhances testing efficiency but also establishes a solid baseline for evaluating VLA models, paving the way for the development of more intelligent and advanced robotic systems.
\end{abstract}


\section{Introduction}

Recent research has demonstrated the application of Large Language Models (LLMs) in various robotic domains~\cite{zeng2023large,wang2024large}, where they are employed to tackle complex tasks that usually require human-like cognitive abilities, such as planning~\cite{huang2022language,kannan2023smart,zhou2024isr}, task comprehension~\cite{ahn2022can,song2023self,lin2023text2motion}, and intention understanding~\cite{huang2022inner,li2022pre,street2024llm}. 
Building on these advancements, a growing number of recent works also employ Vision Language Models (VLMs)~\cite{du2022survey} to enhance robots with the ability to process visual inputs~\cite{zhang2024vision,cui2024survey,long2024robollm}.
It enables robots to interpret their surrounding environments and identify interactable objects, facilitating autonomous decision-making processes necessary for task completion.

\begin{figure}
    \centering
    \includegraphics[trim={2mm 14mm 2mm 2mm},width=0.85\linewidth]{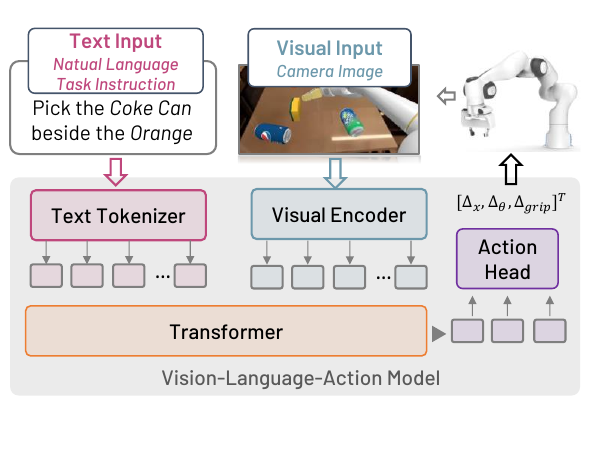}
    \vspace{5pt}
    \caption{The VLA model takes camera images and natural language task instructions as inputs. Using a transformer-based encoding and decoding process, the VLA model directly generates control commands for the robots.}
    \label{fig.VLA}
    \vspace{-15pt}
\end{figure}

This progress has given rise to a new class of end-to-end models known as Vision-Language-Action (VLA) models~\cite{padalkar2023open,ma2024survey}, primarily designed for robotic manipulation tasks~\cite{brohan2022rt,zitkovich2023rt,kim2024openvla}. 
The inputs to the VLA models consist of images captured by cameras and user-provided natural language instructions that describe the desired task. 
The VLA models then directly generate control commands, e.g., the pose of the end-effector, to guide the robotic manipulator in completing the assigned tasks based on the inputs~\cite{brohan2022rt} (see Fig.~\ref{fig.VLA}). 
These large pre-trained VLA models offer a novel approach to robotic control that not only mitigates the need for programming low-level task and motion controllers but also fosters direct interaction between robots and users through natural language instructions. 
This innovation marks a promising advancement toward achieving higher levels of robot intelligence, bringing us closer to the realization of fully autonomous and intelligent robotic systems~\cite{xu2024survey}.

However, the data-driven nature of VLA models introduces several challenges. 
For example, the effectiveness of task execution is highly reliant on the quality of the training data used to develop these models~\cite{padalkar2023open}. 
Moreover, the limited interpretability of VLA models raises concerns about their reliability, robustness, and trustworthiness~\cite{myers2024foundation}.
These challenges underscore the need for a comprehensive platform to test and evaluate the performance of VLA models across a variety of manipulation tasks and scenarios.

\begin{figure*}
    \centering
    \vspace{-10pt}
    \includegraphics[trim={4cm 4.4cm 4cm 2.4cm},width=0.9\linewidth]{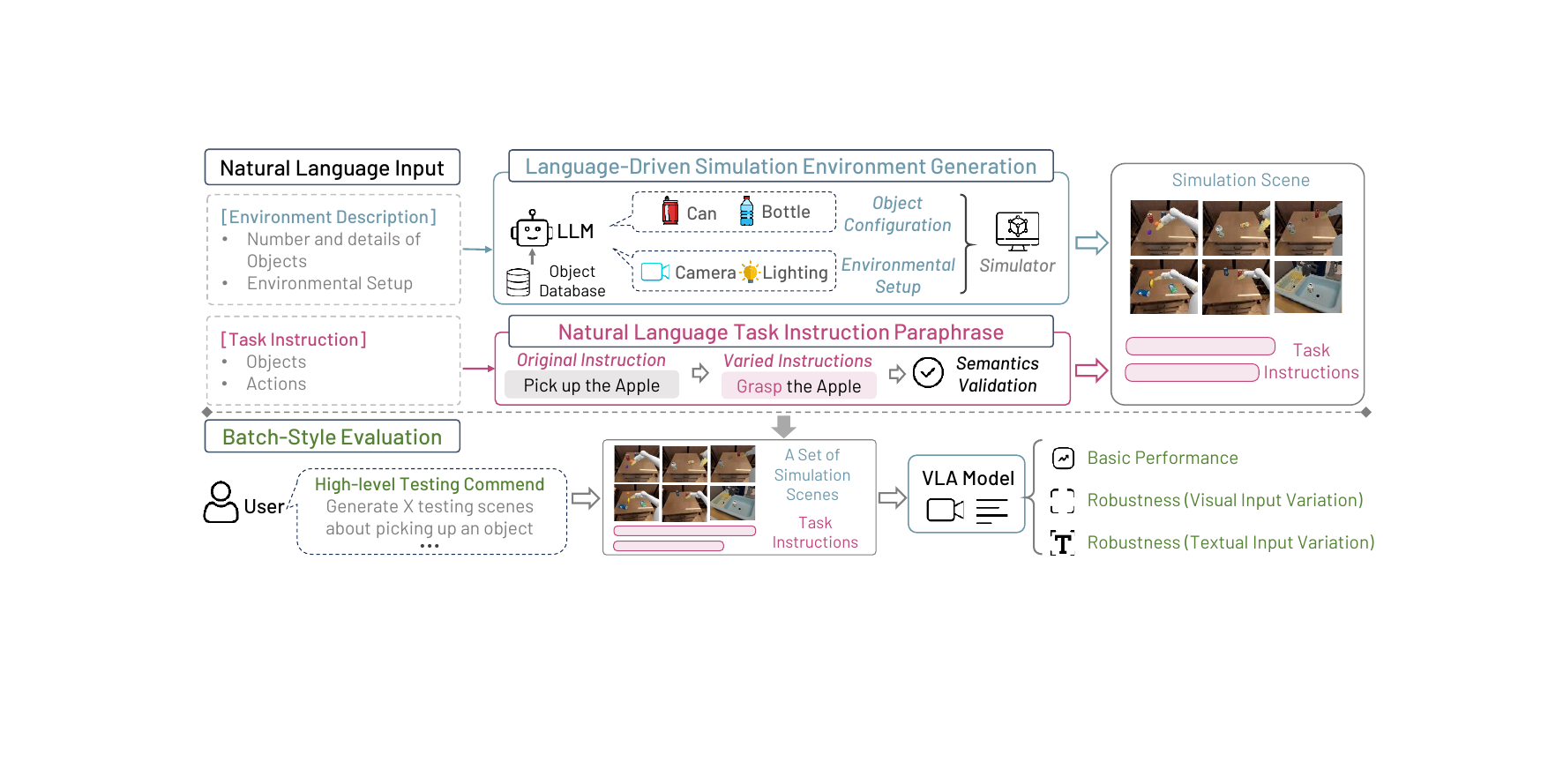}
    \caption{Overview of LADEV. LADEV proposes: (1) Language-driven simulation environment generation; (2) Natural language task instruction paraphrase; (3) Batch-style evaluation. For details about the prompts used in this work, please refer to the preprint version or the supplementary website of this paper. }
    \label{fig.overview}
    \vspace{-15pt}
\end{figure*}

Unfortunately, as an emerging field, related quality assurance methods are still at a very early stage, and there is currently no platform specifically designed to test and evaluate VLA models automatically. This gap in evaluation frameworks highlights the need for reliable tools to measure the performance and robustness of these models. In response to this need,
a simulation platform called SimplerEnv was introduced in ~\cite{li2024evaluating}. 
Built on the SAPIEN simulator~\cite{Xiang_2020_SAPIEN} and the ManiSkill2 benchmark~\cite{gu2023maniskill2}, it includes multiple typical pick-and-place scenarios and various VLA models. 
By generating simulation environments that replicate real-world training conditions, SimplerEnv is able to assess the performance of VLA models in a simulated setting. 
However, modifying the simulation environments in SimplerEnv requires manual adjustments, which can be labor-intensive when testing numerous different environments for comprehensive evaluations. 
Moreover, SimplerEnv alters only the simulated manipulation scenes, e.g., the objects in the environment, which solely affect the visual input to the VLA models. 
The natural language task instruction, i.e., a crucial input component that specifies the manipulation tasks, remains unchanged in SimplerEnv.
To thoroughly test and evaluate VLA models, it is essential not only to efficiently generate a wide range of manipulation scenarios but also to create diverse natural language task instructions. 
These instructions should describe different tasks or express the same task using varying sentence structures and vocabulary to effectively test the language input aspect of the VLA models.

To achieve this, we propose in this work a comprehensive language-driven testing and evaluation platform called \textbf{LADEV}, which is specifically designed for VLA models.
Building on SimplerEnv, we introduce three major advancements in LADEV:
(1) \textit{Language-driven Simulation Environment Generation}: instead of manual adjustments, we introduce an automated mechanism to generate simulation environments based on simple language descriptions of the desired manipulation scenarios. 
Using LLMs, these descriptions are translated into environmental configurations compatible with the simulator for constructing the simulation environment (see Fig.~\ref{fig.overview}). 
To further expand simulation diversity and incorporate a wide variety of objects, we also integrate LADEV with the YCB object dataset~\cite{calli2015benchmarking}, enabling the automatic selection and inclusion of appropriate object models in the simulation based on the given language input. 
(2) \textit{Natural Language Task Instruction Paraphrase}: in addition to generating simulation environments, we also propose a method for paraphrasing natural language task instructions. 
Given an original instruction for the desired manipulation task, we use LLMs to create alternative sentences that convey the same task but with different sentence structures and wording (see Fig.~\ref{fig.overview}).
(3) \textit{Batch-Style Evaluation}: to further streamline the testing and evaluation process, we implement a batch-style generation mechanism capable of creating numerous distinct test environments from a single command input. 
Specifically, we ask an LLM to generate a complete testing script with descriptions of diverse manipulation scenes, which are then passed to the scene generation process to assess the VLA model's performance in each individual scenario.

The contributions of this paper are summarized as follows:
\begin{itemize}[leftmargin=*]
    \item We propose a novel language-driven approach that autonomously generates simulation environments from natural language descriptions of the desired manipulation tasks. 
    This fully automated process greatly improves the efficiency of testing and evaluating VLA models, providing a solid foundation for comprehensive performance assessment.
    \item  We present a paraphrase mechanism that transforms the given natural language task instruction into various forms, enabling a comprehensive assessment of VLA models' ability to handle diverse language inputs. 
    This capability fills a gap in prior evaluations of VLA models, which focused exclusively on simulation environments while neglecting the essential role of language input.
    \item We introduce a batch-style generation approach that is able to construct a diverse range of manipulation scenarios from a single input command. 
    This ``one-line'' testing command enables rigorous large-scale testing and evaluation of VLA models in an efficient way.
    \item Using the proposed LADEV platform, we conduct a thorough and extensive evaluation of multiple state-of-the-art VLA models.
    Specifically, we examined the performance of seven VLA models on four robotic manipulation tasks using over \responseline{4,000} distinct scenes, showcasing their actual capabilities in different scenarios.
\end{itemize}

\section{Related Work}

\subsection{LLM and VLM in Robotics}

In recent research, LLMs have been applied to various robotic tasks, such as decision-making~\cite{ahn2022can,li2022pre,lin2023text2motion} and reasoning~\cite{huang2022inner,zeng2022socratic,song2023self}. 
For instance, \cite{ahn2022can} leverages LLMs' semantic capabilities to process natural language instructions, enabling robots to perform tasks assigned by humans through a value function. 
Similarly, \cite{huang2022inner} utilizes LLMs to evaluate the feasibility of task plans in a dialogue-based format, allowing robots to correct their actions as needed.
Other works have explored using LLMs for task and motion planning~\cite{ding2023task,huang2022language,singh2023progprompt,silver2023generalized,silver2022pddl,liu2023llm}. 
For example, \cite{ding2023task} uses LLMs to guide object rearrangement, improving both autonomy and efficiency. 
Meanwhile, \cite{zhou2024isr} explores the potential of LLMs with a self-refinement mechanism for long-horizon sequential task planning, increasing task success rates compared to a zero-shot LLM approach.
The incorporation of LLMs significantly advances robotic intelligence, enhancing both autonomy and interaction with human users.

Extended from LLMs, an increasing number of studies now have utilized VLMs to equip robotic systems with the ability to process visual inputs~\cite{du2023vision,zhang2024vision}.
One common application of VLMs in robotics is reasoning about the environment and identifying interactable objects~\cite{nair2022r3m,huang2023voxposer,pgvlm2024,song2024vlm,liu2024okrobot}. 
For instance, \cite{huang2023voxposer} combines VLM and LLM to generate 3D affordance and constraint maps that guide robotic manipulation tasks. 
Similarly, \cite{pgvlm2024} proposes a physically grounded VLM to improve the interaction between the robot and the object. 
\cite{song2024vlm} introduces a VLM-based navigation approach for determining the robot's motion in human-centered environments.
The integration of visual processing capabilities further enhances robots' understanding of tasks and environments, opening up the potential for achieving general robotic intelligence~\cite{ma2024survey}.

\subsection{VLA Models in Robotics}

VLA models are end-to-end multi-modality foundation models evolved from VLMs~\cite{li2023vision,stone2023open,ma2024survey}. 
Currently, most VLA models are designed for robotic manipulation tasks, such as pick-and-place and grasping~\cite{padalkar2023open,brohan2023rt, team2024octo, kim2024openvla, li2024llara, brohan2022rt}. 
One of the pioneering works in VLA models is RT-1~\cite{brohan2022rt}, which combines a FiLM EfficientNet and a transformer to learn control policies from 130k real-world robot demonstrations. 
RT-2~\cite{brohan2023rt} advances RT-1 by introducing co-fine-tuning, integrating low-level control policies with high-level task planners to create a more comprehensive robotic system.
Since the release of the Open X-Embodiment dataset~\cite{padalkar2023open}, a series of VLA models have been developed by either training or fine-tuning on this dataset, such as OpenVLA~\cite{kim2024openvla}, Octo~\cite{team2024octo}, and LLaRA~\cite{li2024llara}. 
These models have demonstrated strong performance in their respective training environments, showing great potential for enabling intelligent robotic manipulation using only image and language inputs.

However, ensuring the reliability and robustness of VLA models is challenging, as their performance heavily relies on the quality of the training data~\cite{wang2024large}. 
This necessitates an extensive testing and evaluation platform specifically designed for VLA models.
As previously mentioned, the SimplerEnv, introduced in~\cite{li2024evaluating}, provides valuable simulation environments. 
However, it requires manual adjustments for environment construction and neglects the impact of language inputs.
To overcome these limitations, we therefore propose LADEV in this work, which enables a more efficient, comprehensive, and automated evaluation process for VLA models.

\section{LADEV}
\label{sec.method}

In this section, we introduce details about the proposed LADEV platform. 
First, we describe how LLMs are leveraged to generate simulation environments from natural language descriptions of the desired manipulation scenarios.
Next, we introduce a paraphrase mechanism that alters the given natural language task instructions. 
Lastly, we present a batch-style evaluation method that greatly accelerates the evaluation process with improved efficiency. 
Due to the page limit, detailed information about the prompts used in this work is presented in the pre-print version and the website of this paper: {\href{https://sites.google.com/view/ladev}{https://sites.google.com/view/ladev}}.

\subsection{Language-Driven Simulation Environment Generation}
\label{sec.method_language_scene_generation}


The core concept behind the automated generation of simulated manipulation environments is to convert natural language descriptions into simulator-compatible environmental configurations by leveraging LLMs. 
To achieve this, we use a fixed structure for the natural language description, which includes the following components (see also Fig.~\ref{fig.scene_generation_prompt}):
\begin{itemize}[leftmargin=*]
\item \textit{Number and details of objects}: First, we specify the total number of objects and provide additional details, such as their specific types and poses, to be included in the simulation environment. 
If object details are not needed, this part can be left blank.
\item \textit{Environmental setup}: Then, we describe the environmental setup, including the lighting condition and camera pose. 
If not specified, predefined default values will be used.
\end{itemize}
Using this structured description, we apply a two-step process to separately handle the object configuration and environmental setup during the generation process.


\subsubsection{Object Configuration}
\begin{figure}
    \centering
    \includegraphics[width=0.95\linewidth]{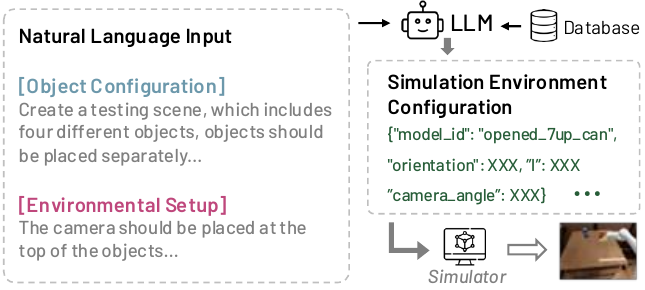}
    \caption{Example of language-driven simulation environment generation. }
    \label{fig.scene_generation_prompt}
    \vspace{-15pt}
\end{figure}

We begin by using the descriptions of the number and details of objects to select appropriate models from LADEV's object model database, which combines the YCB object dataset~\cite{calli2015benchmarking} and the default dataset from SimplerEnv~\cite{li2024evaluating}. 
This is achieved by providing a predefined list of all available objects, along with the natural language description, to the LLM, which then generates a list of object addition operations. 
The length of this list corresponds to the specified number of objects, and each entry represents the addition of an object model to the simulation environment (see Fig.~\ref{fig.scene_generation_prompt}). 
If detailed object specifications are provided, the LLM prioritizes selecting models that best match the criteria. 
For example, if the user requests a \textit{coke can}, LADEV searches for the relevant model and adds it if available. 
When no specific details are given, random objects are selected. 
Similarly, if specific object poses are provided, they are translated into corresponding coordinates. Otherwise, LADEV assigns random values within a predefined range.

\subsubsection{Environmental Setup}
We then prompt the LLM with the description of the environmental setup to configure the simulation parameters. 
In the current version of LADEV, two environmental configurations are considered: the lighting condition and the camera pose. 
If a specific value is provided for the lighting condition, the LLM generates an operation command to adjust the scene's lighting intensity accordingly. 
Similarly, if a camera pose is specified, the LLM generates an operation to move or rotate the camera to match the desired pose, ensuring proper visual inputs for the VLA models. 
If no information is given, predefined default values are applied.

Once the object addition and environmental adjustment operations are generated, we pass them to the simulator to construct the corresponding simulation environment. 
To enhance the accuracy of the LLM’s translations, we employ few-shot in-context learning~\cite{brown2020language} in our prompts. 
This approach also ensures that the LLM outputs are formatted to be compatible with the simulator. 
In LADEV, these operations are specified in JSON configuration formats. 
This language-driven testing automation substantially reduces the time and effort required to construct simulation environments. Moreover, it enables an efficient evaluation of VLA models across diverse manipulation scenarios. To further optimize our workflow, we later propose executing these evaluations in a batch-style process, allowing for more efficient assessment.

\subsection{Natural Language Task Instruction Paraphrase}
\label{sec.method_language_mutation}

To evaluate the performance of VLA models in processing diverse language inputs, we also propose a paraphrase mechanism that generates varied natural language task instructions. 
The paraphrase mechanism consists of two phases: a generation phase and a validation phase (see Fig.~\ref{fig.overview}).


The input to the generation phase is an original task instruction that follows the standard format used in previous works~\cite{brohan2022rt, brohan2023rt, team2024octo, kim2024openvla}, such as using \textit{``pick up apple"} to describe a task involving picking up an apple. 
The goal of the generation phase is to produce a predefined number, $k$, of alternative instructions that convey the same meaning but differ in sentence structure and wording. 
For example, the original instruction \textit{``pick up apple"} could be rephrased as \textit{``grasp apple''}, \textit{``let’s pick the apple''}, or \textit{``can you lift the apple''}, etc. 
This is achieved by prompting an LLM with the original instruction and guidelines for generating alternative sentences. 
The LLM then outputs $k$ variations of the task instructions with distinct wordings and structures. 


After generating a set of candidate sentences, we implement a validation phase to ensure that each sentence accurately retains the same meaning as the original, ensuring the validity of the paraphrased sentences. 
This is achieved by using a \textit{sentence BERT} model~\cite{reimers2019sentence} for similarity checking. 
Specifically, we utilize the \textit{sentence BERT} model to compute embeddings for each language task instruction and measure the semantic similarity between the original ones and the candidate variations. 
If the similarity value exceeds a predefined threshold, the varied task instruction is considered to have the same meaning and is deemed valid. 
These valid instructions are then used to evaluate the VLA models’ performance in handling diverse language inputs.


By utilizing the proposed natural language task instruction paraphrase mechanism, we significantly enhance the diversity of language inputs for VLA models. 
This approach addresses a crucial gap in the overall evaluation, focusing on the previously overlooked aspect of assessing how natural language task instructions impact the performance of VLA models. 

\begin{figure*}
    \centering
    \vspace{-10pt}
    \includegraphics[trim={2cm 3.7cm 2cm 0cm},width=.85\linewidth]{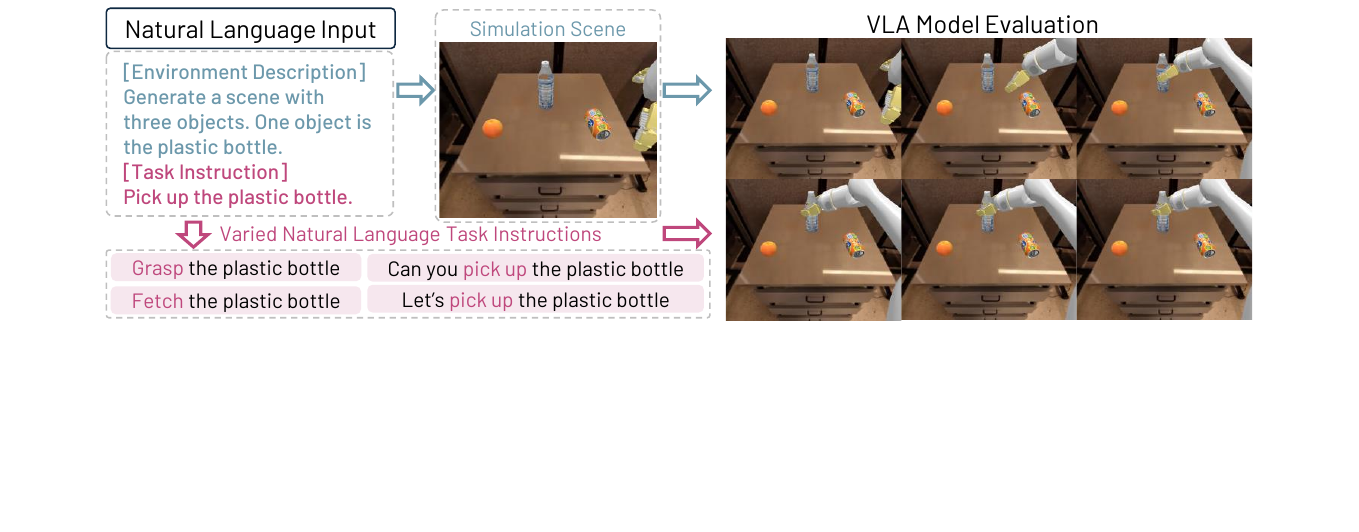}
    \vspace{5pt}
    \caption{Example of simulation scene generation and natural language instruction paraphrase with LADEV. See also the supplementary video for more details.}
    \label{fig.results_concrete_example}
    \vspace{-15pt}
\end{figure*}

\subsection{Batch-Style Evaluation}
\label{sec.method_batch}


Through the methods proposed in Sec.\ref{sec.method_language_scene_generation} and Sec.\ref{sec.method_language_mutation}, we can generate a single manipulation scenario with multiple diverse language task instructions that convey the same task. 
However, for a comprehensive evaluation of the VLA model, it is crucial to test its performance across various scenarios and tasks. 
To expedite this process, we introduce a batch-style evaluation approach that automatically generates a specified number of distinct manipulation scenarios from a single natural language input.

Specifically, we instruct the LLM to automatically generate diverse language inputs for executing both the simulation environment generation and task instruction paraphrase processes (see Fig.~\ref{fig.overview}). 
Suppose the user wishes to create $n$ test scenes with $k$ task instructions per scene. In this case, the LLM is prompted to randomly generate $n$ sets of natural language inputs that specify objects and environmental setups, along with $n$ original task instructions following the standard format. 
Each original task instruction is then used in the paraphrase process to generate $k$ variations, resulting in a total of $n \times k$ language inputs. 
Due to the limited space, the detailed prompt templates and examples used for in-context learning are provided on the website accompanying this paper.



The proposed batch-style evaluation approach enables efficient large-scale testing of VLA models, facilitating a more comprehensive and reliable assessment of their robustness and effectiveness. 
In the next section, we apply this approach to evaluate the performance of multiple state-of-the-art VLA models across various manipulation scenarios.


\section{Experiments}

In this section, we first present the details of our experimental setup. 
Then, we utilize a concrete example to illustrate how LADEV generates a simulation environment from a natural language input and paraphrases the task instruction.
Finally, we present our large-scale experimental results in assessing the performance of multiple state-of-the-art VLA models, which demonstrate the efficiency and effectiveness of the proposed LADEV for VLA models' evaluation.

\subsection{Experimental Setup}

We consider four robotic manipulation tasks in LADEV for evaluating VLA models: (1) \textit{Pick up an object}; (2) \textit{Move object A near object B}; (3) \textit{Put object A on object B}; and (4) \textit{Put object A inside object B}. 
For performance evaluation, we assess the impact of the following factors:
\begin{itemize}[leftmargin=*]
\item \textit{Number of objects}: We first investigated how the number of objects in the simulation scene affects performance. 
More objects introduce additional obstacles, increasing the difficulty for VLA models in correctly identifying the target object. 
To evaluate this, we generated 100 test scenes for each task, with each scene containing 1 to 5 objects. 
Only basic task instructions were used to focus on the effect of the number of objects.

\item \textit{Task instructions}: For each task, we generated 100 test scenes with 1 to 4 randomly selected objects. 
Each scene was evaluated using both the basic task instructions and those paraphrased by LADEV to examine the influence of varied language inputs. For each task, we randomly chose one from ten paraphrased task instructions.

\item \textit{Unseen objects}: The LADEV's object model database is the combination of the SimplerEnv~\cite{li2024evaluating} (18 objects) and the YCB~\cite{calli2015ycb} (65 objects).
Objects in SimplerEnv are generally considered part of the VLA models' training dataset, while objects from the YCB can be regarded as unseen. 
We evaluated the effect of unseen objects by generating two groups of 100 test scenes for each task, randomly including 1 to 4 objects from either the SimplerEnv or the YCB.

\item \textit{Environmental conditions}: To assess the impact of environmental conditions, for each task, we created three sets of 100 test scenes with 1 to 4 objects: one with default lighting and camera settings, one with randomly adjusted lighting conditions, and one with altered camera poses. 
To ensure all objects remain visible and recognizable, only small adjustments were made to lighting conditions \responseline{(increasing or decreasing the lighting intensities by a maximum value of 0.5)} and camera poses \responseline{(a maximum rotation angle of $5^\circ$ and a maximum moving distance of $5$ cm)}.
\end{itemize}

We compare the following state-of-the-art VLA models: RT-1-1k, RT-1-58k, RT-1-400k~\cite{brohan2022rt}, RT-1-X~\cite{padalkar2023open}, Octo-base, Octo-small~\cite{team2024octo}, and OpenVLA-7b~\cite{kim2024openvla}. 
We use GPT-4o as our LLM. 
All experiments are conducted on a server with an AMD 5955WX CPU and two NVIDIA RTX A6000 GPUs. 
For more details about the experimental setup, please refer to the supplementary website of this paper.

\begin{table*}[t]
    \caption{VLA models' performance with different numbers of objects.}
    \label{tab:performance_obstacles}
    \centering
    \scriptsize
    \setlength{\tabcolsep}{3pt}
    \begin{tabular}{lrrrrrrcrrrrrrcrrrrrrcrrrrrr}
        \toprule
        \multirow{2}{*}{VLA Model} & \multicolumn{6}{c}{Pick Up} & & \multicolumn{6}{c}{Move Near} & & \multicolumn{6}{c}{Put On} & & \multicolumn{6}{c}{Put In}\\
        \cmidrule{2-7} \cmidrule{9-14} \cmidrule{16-21} \cmidrule{23-28}
        & \multicolumn{1}{c}{$1$} & \multicolumn{1}{c}{$2$} & \multicolumn{1}{c}{$3$} & \multicolumn{1}{c}{$4$} & \multicolumn{1}{c}{$5$} & \multicolumn{1}{c}{Avg.} && \multicolumn{1}{c}{$1$} & \multicolumn{1}{c}{$2$} & \multicolumn{1}{c}{$3$} & \multicolumn{1}{c}{$4$} & \multicolumn{1}{c}{$5$} & \multicolumn{1}{c}{Avg.} && \multicolumn{1}{c}{$1$} & \multicolumn{1}{c}{$2$} & \multicolumn{1}{c}{$3$} & \multicolumn{1}{c}{$4$} & \multicolumn{1}{c}{$5$} & \multicolumn{1}{c}{Avg.} && \multicolumn{1}{c}{$1$} & \multicolumn{1}{c}{$2$} & \multicolumn{1}{c}{$3$} & \multicolumn{1}{c}{$4$} & \multicolumn{1}{c}{$5$} & \multicolumn{1}{c}{Avg.}\\
        \midrule
        RT-1-1k & 0\% & 1\% & 0\% & 0\% & 0\% & \cellcolor{lightgray}0.2\% && 2\% & 2\% & 1\% & 1\% & 1\% & \cellcolor{lightgray}1.4\% && 0\% & 0\% & 0\% & 0\% & 0\% & \cellcolor{lightgray}0.0\% && 0\% & 0\% & 0\% & 0\% & 0\% & \cellcolor{lightgray}0.0\%  \\
        RT-1-58k & 36\% & 41\% & 27\% & 23\% & 21\% & \cellcolor{lightgray}29.6\% && 11\% & 11\% & 9\% & 10\% & 6\% & \cellcolor{lightgray}9.4\% && 0\% & 0\% & 0\% & 0\% & 0\% & \cellcolor{lightgray}0.0\% && 0\% & 0\% & 0\% & 0\% & 0\% & \cellcolor{lightgray}0.0\%  \\
        RT-1-400k & 44\% & 37\% & 35\% & 33\% & 26\% & \cellcolor{lightgray}35.0\% && 12\% & 14\% & 4\% & 5\% & 3\% & \cellcolor{lightgray}7.6\% && 0\% & 0\% & 0\% & 0\% & 0\% & \cellcolor{lightgray}0.0\% && 0\% & 0\% & 0\% & 1\% & 0\% & \cellcolor{lightgray}0.2\%  \\
        RT-1-X & 26\% & 30\% & 19\% & 16\% & 9\% & \cellcolor{lightgray}20.0\% && 7\% & 12\% & 4\% & 2\% & 4\% & \cellcolor{lightgray}5.8\% && 3\% & 1\% & 1\% & 1\% & 2\% & \cellcolor{lightgray}1.6\% && 0\% & 0\% & 0\% & 1\% & 0\% & \cellcolor{lightgray}0.2\%  \\
        Octo-small & 2\% & 0\% & 1\% & 0\% & 0\% & \cellcolor{lightgray}0.6\% && 3\% & 1\% & 6\% & 0\% & 0\% & \cellcolor{lightgray}2.0\% && 4\% & 5\% & 6\% & 1\% & 2\% & \cellcolor{lightgray}3.6\% && 0\% & 3\% & 0\% & 0\% & 2\% & \cellcolor{lightgray}1.0\%  \\
        Octo-base & 1\% & 0\% & 0\% & 0\% & 0\% & \cellcolor{lightgray}0.2\% && 0\% & 0\% & 0\% & 1\% & 0\% & \cellcolor{lightgray}0.2\% && 0\% & 0\% & 5\% & 0\% & 1\% & \cellcolor{lightgray}1.2\% && 0\% & 1\% & 3\% & 0\% & 1\% & \cellcolor{lightgray}1.0\%  \\
        OpenVLA-7b & 12\% & 7\% & 8\% & 7\% & 2\% & \cellcolor{lightgray}7.2\% && 23\% & 18\% & 12\% & 8\% & 2\% & \cellcolor{lightgray}12.6\% && 1\% & 5\% & 1\% & 1\% & 2\% & \cellcolor{lightgray}2.0\% && 5\% & 1\% & 1\% & 4\% & 0\% & \cellcolor{lightgray}2.2\%  \\
        \bottomrule
    \end{tabular}
    \vspace{-10pt}
\end{table*}

\subsection{Environment Generation and Command Mutation}
We first use a concrete example to demonstrate the processes of simulation environment generation and natural language task instruction paraphrase. 
As shown in Fig.~\ref{fig.results_concrete_example}, the natural language input instructs LADEV to create a simulation environment with three objects, one of which is a plastic bottle. 
After searching the model database, LADEV generates the required environment. 
The basic task instruction, \textit{``pick up the plastic bottle"}, is also paraphrased into four variations by LADEV. 
These task instructions, along with the generated environment, are used to evaluate the performance of VLA models.
This individual process can also be scaled using the proposed batch-style evaluation mechanism, which automatically generates multiple manipulation scenes, each with varied task instructions, to comprehensively test the VLA models' performance.




\subsection{Performance Evaluation of VLA Models}

Using the proposed LADEV, we performed a large-scale evaluation of VLA models by considering the aforementioned factors. 
The success rates for completing the given tasks under various conditions are shown in Table~\ref{tab:performance_obstacles}-Table~\ref{tab:performance_env}. 
We highlight the following key observations:
\begin{itemize}[leftmargin=*]
\item \textit{Number of objects}: As shown in Table~\ref{tab:performance_obstacles}, all VLA models performed best when only one object was present, i.e., only the target object. 
The performance of the VLA models decreased as the number of objects increased. 
When five objects were included, almost all models performed poorly across all test scenes and tasks. 
Model-wise, RT-1-58k, RT-1-400k, and RT-1-X outperformed the other models on the \textit{Pick Up} task, while OpenVLA-7b achieved the highest performance on the \textit{Move Near} task. 
However, for the \textit{Put On} and \textit{Put In} tasks, all VLA models showed poor results, with success rates below 5\%.

\item \textit{Task instructions}: From Table~\ref{tab:performance_nl}, we observed a significant performance drop when using paraphrased instructions compared to the basic ones in the \textit{Pick Up} and \textit{Move Near} tasks. 
However, the performance differences in the \textit{Put On} and \textit{Put In} tasks were marginal, as the basic instructions already resulted in poor performance.

\item \textit{Unseen objects}: As shown in Table~\ref{tab:performance_ycb}, VLA models performed worse with YCB objects compared to SimplerEnv objects. 
For instance, models such as RT-1-58k, RT-1-400k, RT-1-X, and OpenVLA-7b performed relatively well in the \textit{Pick Up} and \textit{Move Near} tasks with SimplerEnv objects.
However, when manipulating YCB objects, their performance dropped by 10\% to 30\%.

\item \textit{Environmental conditions}: Table~\ref{tab:performance_env} reveals that even small changes in environmental conditions could largely affect model performance. 
For lighting condition changes, the RT-1 and RT-1-X models were more adversely affected compared to the others. 
However, for camera pose adjustments, no notable performance differences were observed.
\end{itemize}

\begin{table}[t]
    \caption{Basic task instructions vs. paraphrased (Para.) task instructions.}
    \label{tab:performance_nl}
    \centering
    \scriptsize
    \setlength{\tabcolsep}{3.6pt}
    \begin{tabular}{lrrcrrcrrcrr}
        \toprule
        \multirow{2}{*}{VLA Model} & \multicolumn{2}{c}{Pick Up} & & \multicolumn{2}{c}{Move Near} & & \multicolumn{2}{c}{Put On} & & \multicolumn{2}{c}{Put In}\\
        \cmidrule{2-3} \cmidrule{5-6} \cmidrule{8-9} \cmidrule{11-12}
        & \multicolumn{1}{c}{Basic} & \multicolumn{1}{c}{Para.} & & \multicolumn{1}{c}{Basic} & \multicolumn{1}{c}{Para.} & & \multicolumn{1}{c}{Basic} & \multicolumn{1}{c}{Para.} & & \multicolumn{1}{c}{Basic} & \multicolumn{1}{c}{Para.} \\
        \midrule
        RT-1-1k & 0\% & 2\% && 3\% & 1\% && 0\% & 0\% && 0\% & 0\%  \\
        RT-1-58k & 28\% & 17\% && 12\% & 6\% && 0\% & 1\% && 1\% & 0\%  \\
        RT-1-400k & 36\% & 22\% && 7\% & 3\% && 0\% & 1\% && 0\% & 0\%  \\
        RT-1-X & 20\% & 13\% && 7\% & 4\% && 2\% & 0\% && 1\% & 0\%  \\
        Octo-base & 0\% & 0\% && 2\% & 0\% && 2\% & 3\% && 3\% & 1\%  \\
        Octo-small & 0\% & 1\% && 2\% & 5\% && 4\% & 3\% && 1\% & 1\%  \\
        OpenVLA-7b & 8\% & 7\% && 12\% & 4\% && 2\% & 4\% && 1\% & 2\%  \\
        \bottomrule
    \end{tabular}
    \vspace{-5pt}
\end{table}


\begin{table}[t]
    \caption{Objects from SimplerEnv (SE) vs. objects from YCB.}
    \label{tab:performance_ycb}
    \centering
    \scriptsize
    \setlength{\tabcolsep}{4.3pt}
    \begin{tabular}{lrrcrrcrrcrr}
        \toprule
        \multirow{2}{*}{VLA Model} & \multicolumn{2}{c}{Pick Up} & & \multicolumn{2}{c}{Move Near} & & \multicolumn{2}{c}{Put On} & & \multicolumn{2}{c}{Put In}\\
        \cmidrule{2-3} \cmidrule{5-6} \cmidrule{8-9} \cmidrule{11-12}
        & \multicolumn{1}{c}{SE} & \multicolumn{1}{c}{YCB} & & \multicolumn{1}{c}{SE} & \multicolumn{1}{c}{YCB} & & \multicolumn{1}{c}{SE} & \multicolumn{1}{c}{YCB} & & \multicolumn{1}{c}{SE} & \multicolumn{1}{c}{YCB} \\
        \midrule
        RT-1-1k & 0\% & 0\% && 3\% & 0\% && 0\% & 0\% && 0\% & 1\%  \\
        RT-1-58k & 28\% & 2\% && 12\% & 7\% && 0\% & 0\% && 1\% & 0\%  \\
        RT-1-400k & 36\% & 5\% && 7\% & 3\% && 0\% & 2\% && 0\% & 0\%  \\
        RT-1-X & 20\% & 3\% && 7\% & 0\% && 2\% & 2\% && 1\% & 1\%  \\
        Octo-base & 0\% & 0\% && 2\% & 0\% && 2\% & 1\% && 3\% & 0\%  \\
        Octo-small & 0\% & 0\% && 2\% & 0\% && 4\% & 0\% && 1\% & 3\%  \\
        OpenVLA-7b & 8\% & 0\% && 12\% & 6\% && 2\% & 0\% && 1\% & 0\%  \\
        \bottomrule
    \end{tabular}
    \vspace{-15pt}
\end{table}


\begin{table}[t]
    \caption{Influence of different environmental conditions: default (De.), mutated lighting (Li.), and mutated camera poses (Ca.).}
    \label{tab:performance_env}
    \centering
    \scriptsize
    \setlength{\tabcolsep}{2pt}
    \begin{tabular}{lrrrcrrrcrrrcrrr}
        \toprule
        \multirow{2}{*}{VLA Model} & \multicolumn{3}{c}{Pick Up} & & \multicolumn{3}{c}{Move Near} & & \multicolumn{3}{c}{Put On} & & \multicolumn{3}{c}{Put In}\\
        \cmidrule{2-4} \cmidrule{6-8} \cmidrule{10-12} \cmidrule{14-16}
        & \multicolumn{1}{c}{De.} & \multicolumn{1}{c}{Li.} & \multicolumn{1}{c}{Ca.} & & \multicolumn{1}{c}{De.} & \multicolumn{1}{c}{Li.} & \multicolumn{1}{c}{Ca.} & & \multicolumn{1}{c}{De.} & \multicolumn{1}{c}{Li.} & \multicolumn{1}{c}{Ca.} & & \multicolumn{1}{c}{De.} & \multicolumn{1}{c}{Li.} & \multicolumn{1}{c}{Ca.} \\
        \midrule
        RT-1-1k & 0\% & 0\% & 1\% && 3\% & 3\% & 6\% && 0\% & 0\% & 2\% && 0\% & 0\% & 0\%  \\
        RT-1-58k & 28\% & 23\% & 31\% && 12\% & 12\% & 11\% && 0\% & 0\% & 0\% && 1\% & 1\% & 0\%  \\
        RT-1-400k & 36\% & 20\% & 30\% && 7\% & 6\% & 8\% && 0\% & 0\% & 0\% && 0\% & 0\% & 0\%  \\
        RT-1-X & 20\% & 9\% & 14\% && 7\% & 8\% & 7\% && 2\% & 2\% & 1\% && 1\% & 1\% & 1\%  \\
        Octo-base & 0\% & 0\% & 1\% && 2\% & 2\% & 3\% && 2\% & 2\% & 3\% && 3\% & 3\% & 2\%  \\
        Octo-small & 0\% & 1\% & 0\% && 2\% & 2\% & 2\% && 4\% & 4\% & 2\% && 1\% & 1\% & 2\%  \\
        OpenVLA-7b & 8\% & 12\% & 14\% && 12\% & 12\% & 15\% && 2\% & 2\% & 1\% && 1\% & 1\% & 1\%  \\
        \bottomrule
    \end{tabular}
    \vspace{-15pt}
\end{table}

\section{Discussion}

\subsection{Pros and Cons of VLA Models}

By combining visual, linguistic, and action-based information, VLA models offer several advantages to robotic systems. 
For instance, they enable robots to better interpret their surroundings and execute tasks using natural language instructions, reducing the dependence on hardcoded or structured inputs. 
This leads to more intuitive human-robot communication, enhancing interaction flexibility while boosting the robots' autonomy and intelligence. 
However, VLA models also face multiple challenges. 
For example, training these models requires large, multi-modal datasets and significant computational resources. 
Unfortunately, high-quality datasets that align visual inputs, language descriptions, and corresponding actions are scarce. 
Moreover, our experiments show that current VLA models still struggle with even simple manipulation tasks under varied conditions, highlighting the need for deeper exploration in this area to achieve better performance.

\subsection{Limitations and Future Work}

One limitation of the current version of LADEV is that it only considers four manipulation tasks. 
This is primarily due to the fact that state-of-the-art VLA models are still only trained for simple tasks. 
Another drawback is that our evaluations are conducted solely in simulations, as performing comprehensive real-world experiments is difficult and resource-intensive. 
To reduce the gap between simulation and reality, we adopt the same approach as SimplerEnv by using real-world images as backgrounds for the visual inputs to the VLA models. 
However, further research is needed to better minimize the simulation-to-reality gap and develop more efficient methods for assessing the performance of VLA models in real-world conditions.

For future work, we plan to expand the LADEV platform by incorporating more object models and manipulation scenarios, which would greatly increase its diversity and effectiveness. 
Another potential direction is to compare the performance of VLA models in both simulation and real-world environments across several representative tasks, serving as an indicator of their reliability in practical applications. 
However, how to select the most appropriate and representative tasks will require further in-depth research.

\section{Conclusion}

In this work, we propose LADEV, a testing and evaluation platform for VLA models in robotic manipulation tasks. 
By introducing a language-driven framework, we efficiently generate simulation environments from simple natural language inputs, mitigating the need for manual adjustments. 
To assess the impact of language instructions on VLA models, we also present a task instruction paraphrase approach that automatically generates diverse sentences to enrich the language input.
Moreover, to further improve the evaluation efficiency, we develop a batch-style mechanism that creates multiple testing scenarios from a single command, enabling a comprehensive and streamlined assessment of VLA models' performance. 
Our platform notably improves the evaluation process and establishes a strong baseline for advancing VLA models, paving the way for more intelligent robotic systems with enhanced autonomy and decision-making capabilities.

\bibliographystyle{IEEEtran}
\bibliography{references}

\end{document}